\documentclass{article}
\usepackage{arxiv}

\usepackage[utf8]{inputenc} 
\usepackage[T1]{fontenc}    
\usepackage{hyperref}       
\usepackage{url}            
\usepackage{booktabs}       
\usepackage{amsfonts}       
\usepackage{nicefrac}       
\usepackage{microtype}      
\usepackage{lipsum}		
\usepackage{graphicx}
\usepackage{doi}
\usepackage{amsmath} 

\title{PuYun: Medium-Range Global Weather Forecasting Using Large Kernel Attention Convolutional Networks}

\author{Shengchen Zhu\thanks{First author: \texttt{zhushengchen@metac-inc.com}} \and \textbf{Yiming Chen} \and \textbf{Peiying Yu} \and \textbf{Xiang Qu} \and \textbf{Yuxiao Zhou} \and \textbf{Yiming Ma} \vspace{2mm}
 \and \textbf{Zhizhan Zhao} \and \textbf{Yukai Liu} \and \textbf{Hao Mi} \and \textbf{Bin Wang}\thanks{Corresponding author: \texttt{wangbin@metac-inc.com}}}

\date{\vspace{-5mm}\textbf{MetaCarbon} \\August 30, 2024}

\renewcommand{\headeright}{A Preprint}
\renewcommand{\undertitle}{A Preprint}
\renewcommand{\shorttitle}{PuYun: Medium-Range Global Weather Forecasting}

\hypersetup{
pdftitle={PuYun: Medium-Range Global Weather Forecasting Using Large Kernel Attention Convolutional Networks},
pdfauthor={Shengchen Zhu, Yiming Chen and etc.},
pdfkeywords={Weather Forecasting, Large Kernel Attention, Autoregressive Model, Cascading},
}

\begin{document}
\maketitle

\begin{abstract}

Accurate weather forecasting is essential for understanding and mitigating weather-related impacts. In this paper, we present PuYun, an autoregressive cascade model that leverages large kernel attention convolutional networks. The model's design inherently supports extended weather prediction horizons while broadening the effective receptive field. The integration of large kernel attention mechanisms within the convolutional layers enhances the model's capacity to capture fine-grained spatial details, thereby improving its predictive accuracy for meteorological phenomena.

We introduce PuYun, comprising PuYun-Short for 0-5 day forecasts and PuYun-Medium for 5-10 day predictions. This approach enhances the accuracy of 10-day weather forecasting. Through evaluation, we demonstrate that PuYun-Short alone surpasses the performance of both GraphCast and FuXi-Short in generating accurate 10-day forecasts. Specifically, on the 10th day, PuYun-Short reduces the RMSE for Z500 to 720 m²/s², compared to 732 m²/s² for GraphCast and 740 m²/s² for FuXi-Short. Additionally, the RMSE for T2M is reduced to 2.60 K, compared to 2.63 K for GraphCast and 2.65 K for FuXi-Short. Furthermore, when employing a cascaded approach by integrating PuYun-Short and PuYun-Medium, our method achieves superior results compared to the combined performance of FuXi-Short and FuXi-Medium. On the 10th day, the RMSE for Z500 is further reduced to 638 m²/s², compared to 641 m²/s² for FuXi. These findings underscore the effectiveness of our model ensemble in advancing medium-range weather prediction. Our training code and model will be open-sourced.
\end{abstract}


\section{Introduction}
\label{sec:intro}

Accurate weather forecasting bolsters emergency management and mitigation strategies, thus safeguarding lives. It also helps mitigate substantial financial losses incurred by severe weather incidents and yields significant economic benefits\cite{thompson1955economic, thompson1962economic, palmer2000predicting, simmons2002some, 2009300, kleiber2011geostatistical}. Traditional methods of weather forecasting rely on numerical weather prediction (NWP), which uses numerical techniques to analyze the atmospheric state as a grid and determine transitions between states\cite{harper200750th, hu2023progress}. NWP is an essential tool for providing accurate and reliable meteorological data, with positive impacts across various societal sectors \cite{bauer2015quiet, roulston2006laboratory}.
However, due to the highly non-linear and variable nature of weather phenomena\cite{krishnamurthy2019predictability}, NWP based forecasts still struggle with accurately predicting specific scenarios (e.g. monsoons, cyclones and etc.)\cite{brunner2021challenges, krishna2005advancing, wang2005fundamental}. Furthermore, the computational expenditure associated with these predictions is exorbitantly high \cite{bauer2020ecmwf, zangl2015icon}.

In contrast to NWP models, machine learning (ML) methods exhibit the advantage of not relying on a prior understanding of intricate physical processes. Instead, ML models autonomously discern and leverage patterns inherent in data \cite{lecun2015deep, carleo2019machine, chen2023causality, schultz2021can, camporeale2019challenge}. Recent strides, exemplified by methods like FourCastNet\cite{pathak2022fourcastnet}, Pangu\cite{bi2023accurate}, Graphcast\cite{lam2023learning}, and FuXi\cite{chen2023fuxi}, have achieved commendable results, thereby paving the way for further advancements in this field. However, it is noteworthy that these models predominantly operated at a spatial resolution of 0.25°\cite{pathak2022fourcastnet, bi2023accurate, lam2023learning}. An element corresponds to an area of approximately ${25 \times 25}$ square kilometers. This limitation curtails their efficacy in applications demanding more refined forecasts. Sectors such as renewable energy generation or highly localized extreme weather prediction necessitate forecasts of superior spatial precision\cite{kurinchi2021wisosuper, camal2022towards, schicker2023weather, schicker2023post}. This discrepancy underscores the need for advancements in spatial resolution to enhance the applicability of ML models in domains where granular forecasts are imperative. 

Our work, PuYun\footnote{PuYun, in Chinese, refers to beautiful clouds. In ancient China, people believed that clouds could reflect future weather conditions, serving as an important basis for predicting the weather.}, aims to utilize convolution for describing local weather interactions and the more accurate long-term weather forecasting benefits brought by large kernel attention mechanisms. The advent of Vision Transformers (ViT) \cite{dosovitskiy2020image,liu2021swin}, they have rapidly surpassed convolutional networks (ConvNets) \cite{liu2022convnet}, asserting dominance in the realm of computer vision\cite{khan2022transformers}.  In contrast to previous works, PuYun re-embraces the merits of ConvNets, adopting a fully convolutional network (FCN) with large kernel attention (LKA) as its principal framework. Unlike ViT-based approaches that may struggle with resolution expansion, FCN seamlessly accommodate the augmentation of resolution. This attribute proves especially beneficial in contexts where detailed and high-resolution information holds paramount importance\cite{camal2022towards, kurinchi2021wisosuper, schicker2023weather, schicker2023post}. 

Deviating from the conventional strategy in ConvNets, which stacks deep networks with small convolution kernels, PuYun employs LKA convolutions and integrates attention mechanisms with a substantial window size. Given the spatial locality inherent in atmospheric systems, where the weather in a specific region can be influenced by nearby areas, the large kernel attention mechanisms collectively establish a significantly effective receptive field. This enables PuYun to capture spatial locality and enhance predictive capabilities for meteorological phenomena. Furthermore, our PuYun model incorporates a self-regression and cascading structure, facilitating long-term weather forecast predictions. Notably, a distinctive feature of PuYun is its ability to serve as a pretrained model, fine-tuned at higher resolutions (0.1°), thereby enabling higher-resolution forecasts, rendering it remarkably flexible and versatile. 

First, we trained a model to predict the next step, referred to as the single-step model. Building on this model, we applied dynamic step autoregressive training, where the single-step model was fine-tuned with different GPUs assigned varying autoregressive steps randomly sampled from 2 to 12. This process resulted in the PuYun-Short and PuYun-Medium models, which are cascaded to forecast weather for the next 10 days. Using only the PuYun-Short model, the forecast accuracy on the 10th day surpasses that of GraphCast and FuXi-Short. When predictions are made using the cascaded approach, PuYun also outperforms FuXi in the 10-day forecast.

\section{Method}
\begin{figure*}[!t]
  \centering \includegraphics[width=\textwidth]{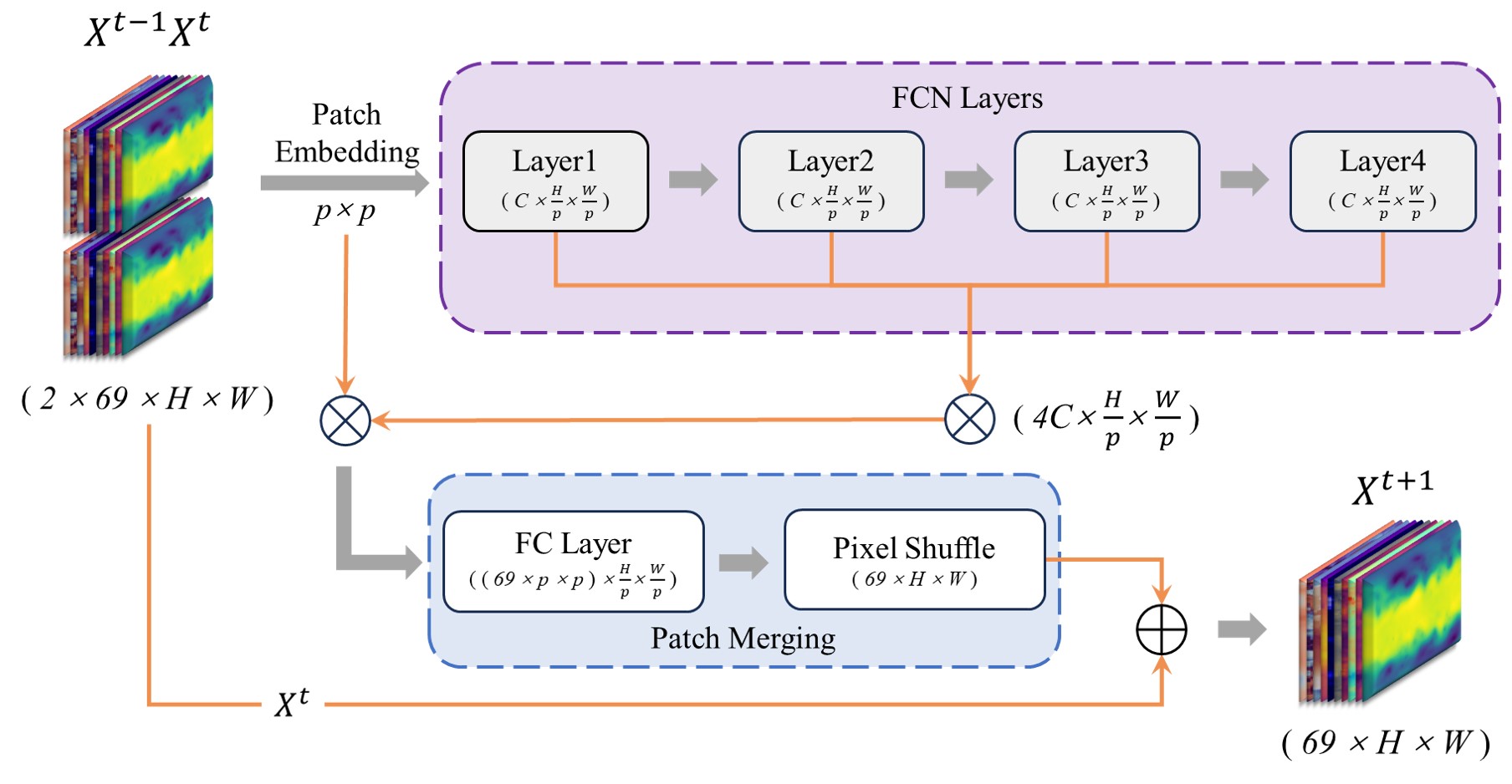}
  \caption{\textbf{Overview of our proposed PuYun model.} It consists of three core components: patch embedding, LKA-FCN layers (the dashed box above) and patch merging (the dashed box below.). The symbol $\otimes$ means concatenation and the $\oplus$ means addition. Skip connections are depicted by thin arrowed lines.} 
  \label{fig:architecture}
\end{figure*}

In this section, we introduce a new medium-range global weather forecasting model using LKA ConvNets called "PuYun". An overview of our method is given in Fig. \ref{fig:architecture}. PuYun leverages the innovative LKA-FCN, offering the unique capability of resolution expansion, a feature absent in ViTs.
The input data combines both atmospheric and surface variables, forming a $2 \times 69 \times 721 \times 1440$ tensor. Here, 2, 69, 721, and 1440 correspond to the two preceding time steps ($t-1$ and $t$), the total number of variables ($C$), latitude ($H$) and longitude ($W$) grid points, respectively. This combined data is then fed into the PuYun model. Subsequent sections delve into the specifics of each component within the PuYun model.

\subsection{PuYun Model}

\subsubsection{Patch embedding}
The technique of patch embedding, commonly used in computer vision, is applied for dimensionality reduction. This entails transforming the data from its original space into a latent space of $C$ dimensions. As depicted in Fig. \ref{fig:architecture} and referred to as Patch Embedding, this operation reduces the data dimensionality to $C \times 90 \times 180$, with a patch size of $8 \times 8$ and a stride of $8 \times 8$. This optimization enhances computational and memory efficiency for self-attention calculations.

\subsubsection{LKA-FCN layers}
\begin{figure}[h]
    \centering
    \includegraphics[width=0.3\textwidth]{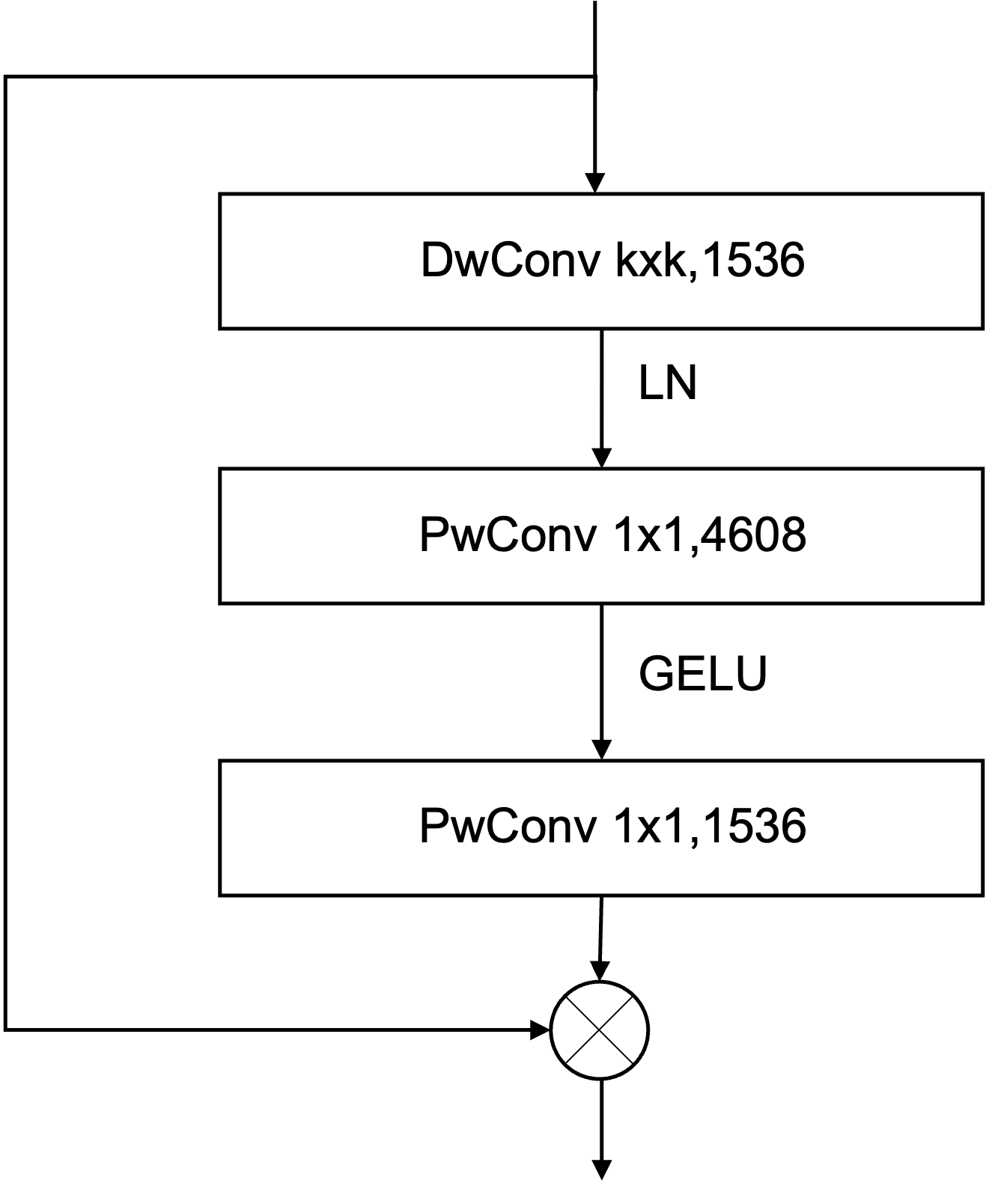}
    \caption{\textbf{Basic block in LKA-FCN Layers.} Each LKA-FCN layer consists of a series of stacked blocks. Each block has a dimension of 1536 with a kernel size of {K}. The symbol $\otimes$ means Hadamard product.}
    \label{fig:lka}
\end{figure}

Instances of LKA-FCN layer are stacked to constitute the foundational core of the PuYun model, each contributing to the hierarchical feature extraction process. Specifically, the PuYun model incorporates 4 LKA-FCN layers, each comprising 12 blocks. The structure of the block is illustrated in Fig. \ref{fig:lka}.

These LKA-FCN layers are instrumental in discerning intricate spatial patterns, where each block functions as a LKA block characterized by a kernel size of {K}. The strategic organization of LKA-FCN layers empowers the model to iteratively enhance its comprehension of the input, extracting diverse features and refining its ability to decipher complex spatial relationships.

\vspace{-3mm}

Following patch embedding, the processed data is fed into LKA-FCN layers. The layer-normalized outputs from these layers are subsequently concatenated along the channel dimension, resulting in a cohesive data with dimensions of $4C \times 90 \times 180$. This concatenation operation adeptly consolidates hierarchical information gleaned from diverse features, thereby augmenting the overall feature representation. To further enrich this integrated information, a skip connection is introduced, concatenating the output from the patch embedding with the result of the initial concatenation operation. This strategic inclusion enhances the model's ability to capture nuanced features and foster comprehensive understanding throughout the processing stages.

\subsubsection{Patch merging}
During the patch merging process, the data fed into a fully-connected layer, refining the features from the $C$ channels to dimensions of $69 \times 8 \times 8$. Following this transformation, the features undergo a pixel shuffle operation designed to upscale the spatial resolution of the data to $720 \times 1440$, subsequently resized to $721 \times 1440$. As the final step, the input $X^t$ is added, culminating in the production of the ultimate result.

\subsubsection{Autoregressive Forecasting}
PuYun refines input from two weather states, $\left(X^{t-1}, X^t\right)$, representing the current time, $t$, and the immediately preceding time, $t-1$, to forecast the weather state at the subsequent time step. The time step considered in this model is 6 hours. Formally, it is expressed as follows:
\begin{equation}
\label{PuYun}
\hat{X}^{t+1}=\mathit{PuYun}\left(X^{t-1}, X^t\right) %
\end{equation}
To generate a $T$-step forecast, $\hat{X}^{t+1: t+T}=\left(\hat{X}^{t+1}, \ldots, \hat{X}^{t+T}\right)$, PuYun employs Equation (\ref{PuYun}) iteratively in an autoregressive manner. It feeds its own predictions back as input to forecast subsequent steps. For example, to predict step $t+1$, the input is $\left(X^{t-1}, {X}^t\right)$. To predict step $t+2$, the input is $\left(X^{t}, \hat{X}^{t+1}\right)$.

\begin{figure}[h]
    \centering
    \includegraphics[width=1.0\textwidth]{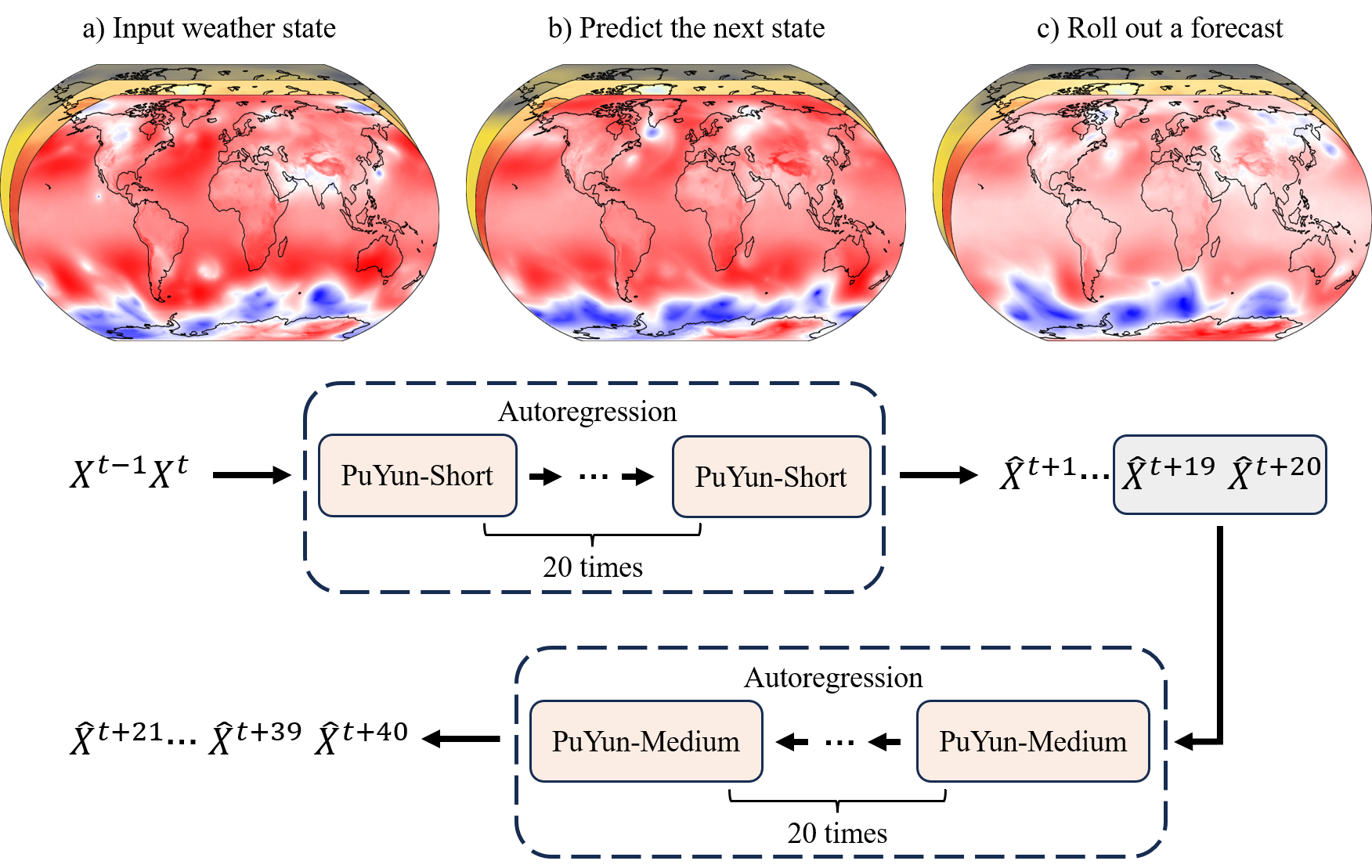}
    \caption{\textbf{Pipeline of PuYun for global weather forecasting.} PuYun-Short generates forecasts for 0-5 days using autoregression, while PuYun-Medium generates forecasts for 5-10 days in a similar manner with the outputs of PuYun-Short.}
    \label{fig:pipline}
\end{figure}

The pipeline for PuYun's 10-day forecast, with predictions made every 6 hours, is depicted in Fig. \ref{fig:pipline}.

\subsection{Loss Function}
PuYun undergoes pre-training with the goal of minimizing a loss function for a single time step against ERA5 targets, employing the gradient descent optimization technique. The specific loss function utilized for single-step training is mean absolute error (MAE), defined as:
\begin{equation}
L_{MAE} = \frac{1}{C \times H \times W} \sum_{c=1}^C \sum_{i=1}^H \sum_{j=1}^W a_i \left| \hat{X}_{c, i, j}^{t+1} - X_{c, i, j}^{t+1} \right|
\end{equation}

where $C$, $H$, and $W$ represent the number of channels and the grid points in the latitude and longitude directions, respectively. The variables $c$, $i$, and $j$ denote the indices for channels, latitude coordinates, and longitude coordinates, respectively.
The $\hat{X}_{c, i, j}^{t+1}$ and $X_{c, i, j}^{t+1}$ correspond to the predicted and ground truth values for a specific variable and locations (latitude and longitude coordinates) at time step $t+1$.
$a_i$ represents the weight at latitude $i$, with the value of $a_i$ decreasing as latitude increases. 
The $L_{MAE}$ loss is then averaged across all grid points and variables.

For autoregressive training, which extends to multiple time steps, the mean squared error (MSE) loss function was used.

\subsection{Datasets}
\label{datasets}
Our work utilizes ECMWF ReAnalysis v5 (ERA5) \cite{hersbach2020era5}. The ERA5 dataset, produced by the ECMWF, represents the fifth generation of ECMWF reanalysis data. Covering from 1940 to the present, it offers comprehensive information on Earth's climate and weather conditions. 
The data is provided at a 0.25° latitude-longitude resolution and spans 37 vertical pressure levels, ranging from 1000 hPa to 1 hPa, making it suitable for various applications in climate research, weather forecasting, and environmental monitoring.  We sample 00/06/12/18 time slots from the original ERA5 dataset.

To train and test the PuYun model, we utilize 40 years of historical weather data (1979-2018) from ERA5 reanalysis archive. Specifically, data from 1979 to 2016 is used for training, data from 2017 for validation, and data from 2018 for testing. For each pressure level, five variables are considered: geopotential height (Z), relative humidity (R), temperature (T), and the u and v components of wind speed (U, V). Additionally, four surface-level variables are included: 2-meter temperature (2T), u-component and v-component of 10m wind speed (10U, 10V), and mean sea level pressure (MSL).

During the training phase on ERA5, our observations revealed that the inclusion or exclusion of precipitation has negligible effects on the ultimate accuracy of other meteorological variables. As a result, we made the decision to strategically remove precipitation from our model.

\subsection{Implementation Details}
The PuYun model is constructed utilizing the PyTorch framework \cite{2017Automatic}. The pretraining of the model necessitates approximately 96 hours, employing a cluster comprising 32 Nvidia A100 GPUs. The training process employs a batch size of 1 on each GPU and ultimately achieves a total batch size of 32 (we observed that the batch size significantly influences the accuracy. Even with just 8 A100 GPUs, employing gradient accumulation to raise the batch size to 32 allows us to attain comparable outcomes). This process involves 120,000 iterations. We normalized the ERA5 data and compressed it into the Zarr format, resulting in a final data size of 4.3T. The I/O becomes the main bottleneck in the early phase of distributed training in our clusters. We used a data cache strategy to cache a copy of the data on the disk of each GPU machine along the training process, which reduces the entire training process from 492 hours to 96 hours.

The AdamW optimizer \cite{2014Adam} is employed with parameters $\beta_1=0.9$ and $\beta_2=0.95$, utilizing an initial learning rate of $1 \times 10^{-3}$ and a weight decay coefficient of 0.1. To mitigate overfitting, Scheduled DropPath \cite{2016FractalNet} is implemented with a dropping ratio of 0.2. For efficient memory management during model training, Fully-Sharded Data Parallel (FSDP) \cite{zhao2023pytorch}, bfloat16 floating-point precision, and gradient check-pointing \cite{chen2016training} are incorporated.

\subsection{Training Procedure}
\label{training}
After training the single-step model, we fine-tuned it to obtain two refined models: PuYun-Short, and PuYun-Medium. The fine-tuning process employed dynamic step autoregressive training, where different GPUs were assigned varying autoregressive steps sampled randomly from 2 to 12. After pre-training, the PuYun-Short model underwent training on data from 1979 to 2016 for 10,000 steps. Subsequently, the PuYun-Short model generated a five-day forecast for the period 2010-2017, serving as input for the PuYun-Medium model. The PuYun-Medium model then underwent another training for 10,000 steps. This finetuning process is specifically designed to achieve optimal performance in generating 6-hourly forecasts for up to 10 days. During finetuning, the learning rate is fixed at $3 \times 10^{-7}$, and the batch size is 32.

\section{Model Evaluation}
To evaluate the performance of the PuYun model, it is imperative to conduct a comparative analysis with state-of-the-art NWP and ML-bsaed models. In alignment with previous methodologies\cite{lam2023learning,bi2023accurate}, we have selected 00z and 12z as the initial forecast times. For deterministic forecasts, the Root Mean Square Error (RMSE) and Anomaly Correlation Coefficient (ACC) are utilized as the primary evaluation metrics.

RMSE represents the latitude-weighted Root Mean Square Error. Given the prediction result \(\hat{x}_{c,i}^{t_0+\tau}\) and its target (ground truth) \(x_{c,i}^{t_0+\tau}\), the RMSE is defined as follows:
\begin{equation}
\mathcal{L}^{c,\tau}_{\text{RMSE}} = \frac{1}{|T|} \sum_{t_0 \in T} \sqrt{\frac{1}{|G_{0.25}|} \sum_{i \in G_{0.25}} a_i \left( \hat{x}^{t_0+\tau}_{c,i} - x^{t_0+\tau}_{c,i} \right)^2} 
\end{equation}
Here, \(c\) indicates the index for channels, which could be either surface variables or atmospheric variables at certain pressure levels. \(i \in G_{0.25}\) are the location (latitude and longitude) coordinates in the grid. \( a_i \) is the area of the latitude-longitude grid cell (normalized to unit mean over the grid) which varies with latitude.

ACC is the Latitude-weighted Anomaly Correlation Coefficient that evaluates the performance of dynamical models by comparing their predictions of anomalies (departures from the long-term averaged climatology) to observed anomalies.
\begin{equation}
\mathcal{L}^{c,\tau}_{\text{ACC}} = \frac{1}{|T|} \sum_{t_0 \in T} \frac{\sum_{\substack{i \in G_{0.25}}} (\hat{x}'_{c,i} \cdot x'_{c,i})}{\sqrt{\sum_{\substack{i \in G_{0.25}}} (\hat{x}'_{c,i})^2} \cdot \sqrt{\sum_{\substack{i \in G_{0.25}}} (x'_{c,i})^2}}
\end{equation}
where \( \hat{x}'_{c,i} = a_i (\hat{x}^{t_0+\tau}_{c,i} - m^{t_0+\tau}_{c,i}) \) and \( x'_{c,i} = a_i (x^{t_0+\tau}_{c,i} - m^{t_0+\tau}_{c,i}) \). In this context, \(m_{c,i}^{t_0+\tau}\) is the climatological mean over the day-of-year containing the validity time \(t_0 + \tau\) for a given weather variable \(c\) at longitude \(w\) and latitude \(h\). It is averaged from daily data using ERA5 data from 1993 to 2016.

\section{Experimental Results}
PuYun can predict 69 meteorological variables. We selected several representative meteorological variables for comparison, which are as follows.
\begin{itemize}
\item[$\bullet$] Z500: Geopotential at 500 hPa.
\item[$\bullet$] T500: Temperature at 500 hPa.
\item[$\bullet$] U500, V500: U and V components of wind at 500 hPa.
\item[$\bullet$] 2T: 2 meter temperature.
\item[$\bullet$] 10U, 10V: U and V components of wind at 10 meter.
\item[$\bullet$] MSL: Mean sea level pressure.
\end{itemize}

2T, 10U, 10V and MSL are surface meteorological variables that are directly perceptible to humans. The 500hPa level, representing atmospheric pressure at 500 millibars, holds significance in meteorology due to its typical location in the mid-level atmosphere, approximately halfway up in the atmosphere's vertical extent. This positioning enables meteorological variables at the 500hPa level to provide crucial insights into large-scale weather systems.

We compared PuYun with leading ML based models Pangu, GraphCast and FuXi.

\subsection{Quantitative Comparison}
To evaluate the performance of PuYun, we utilize data from the year 2018 and selected two daily initialization times (00:00 UTC and 12:00 UTC) to generate 10-day forecasts with 6-hour intervals.

\begin{figure}[h]
    \centering
    \includegraphics[width=0.92\textwidth]{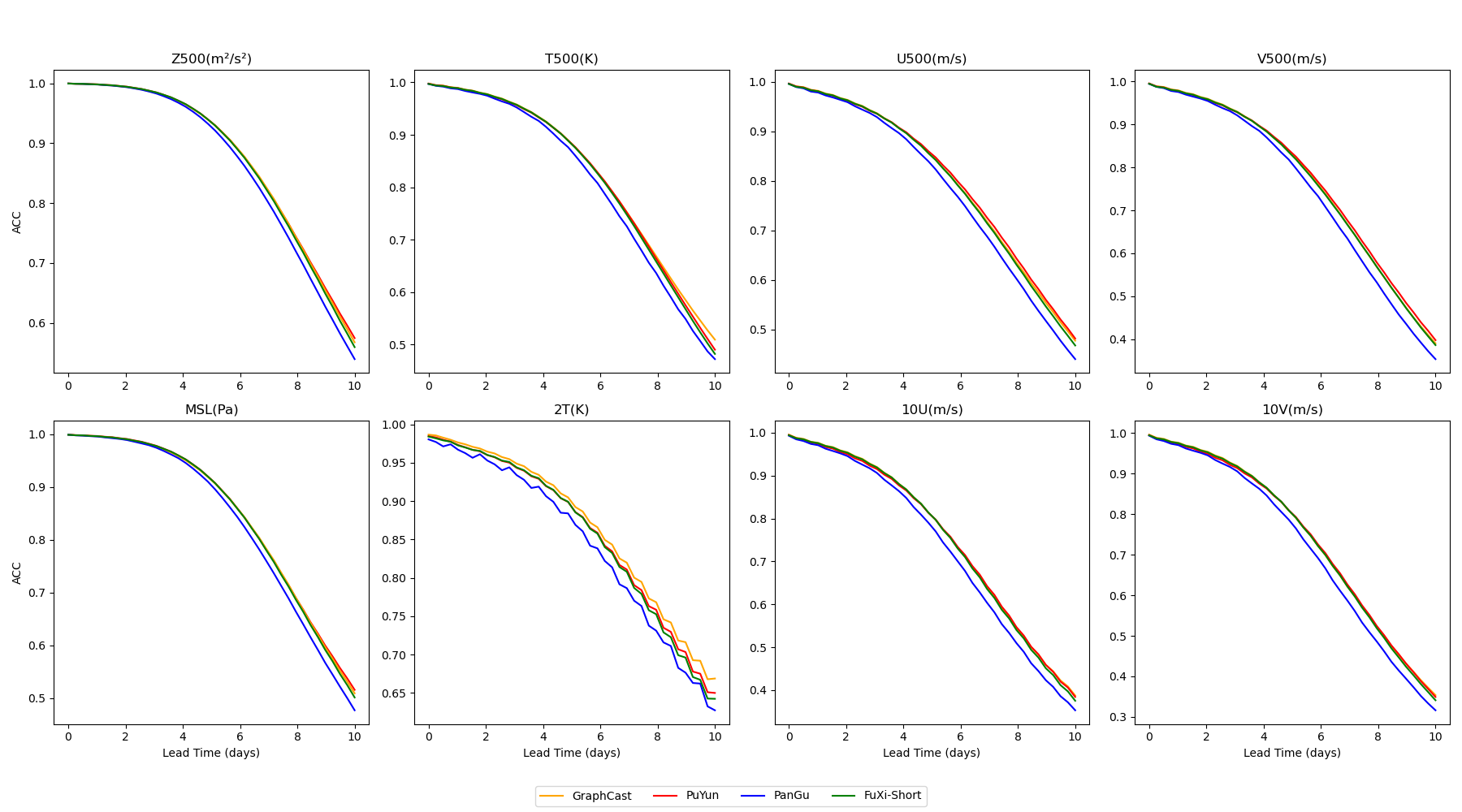}
    \caption{Forecast ACC comparison among Pangu, GraphCast, FuXi-Short and PuYun-Short for an array of meteorological elements over a 10-day forecast period. Higher numerical values indicate better performance.}
    \label{fig:single_acc}

    \includegraphics[width=0.92\textwidth]{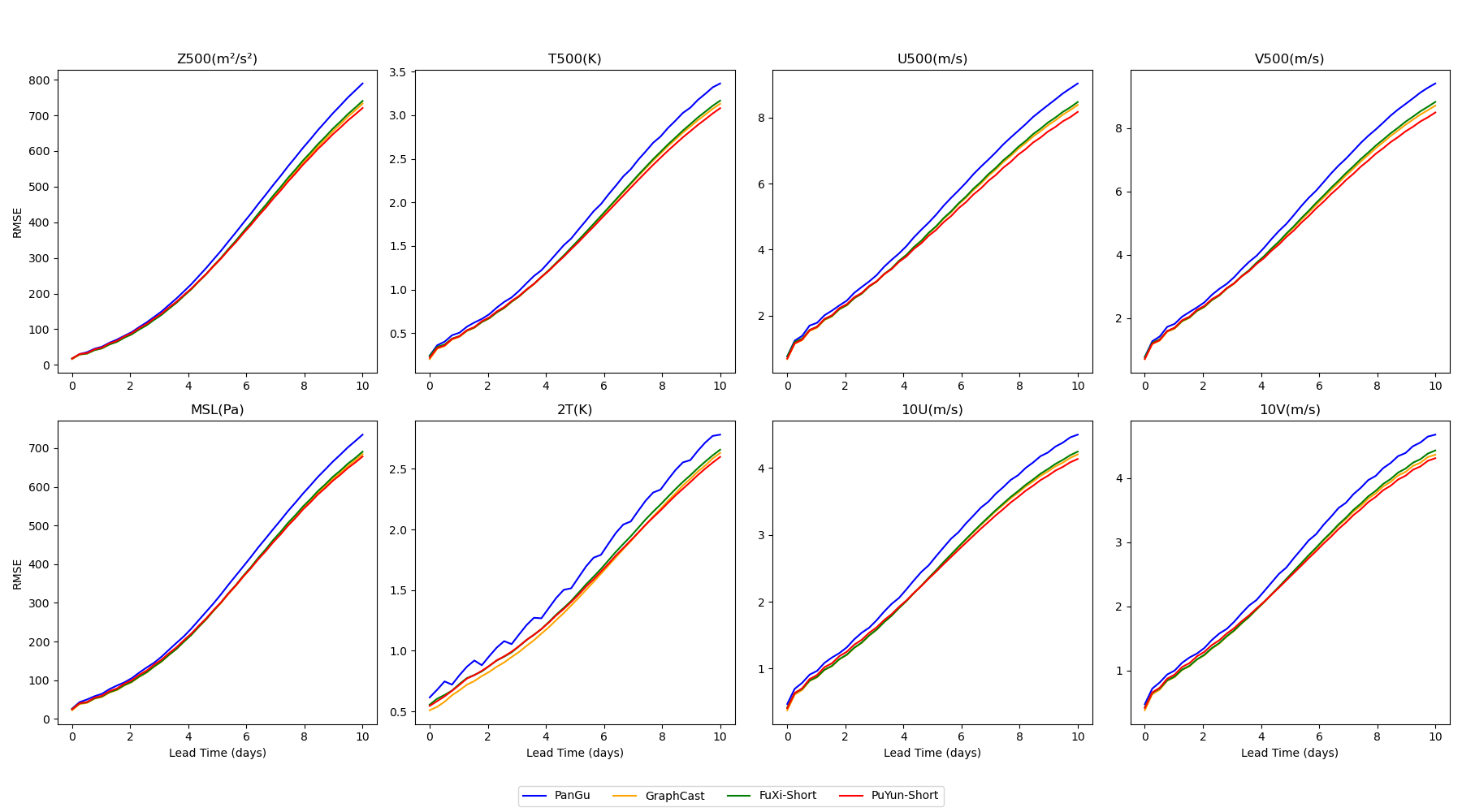}
    \caption{Forecast RMSE Comparison among Pangu, GraphCast, FuXi-Short and PuYun-Short for an array of meteorological elements over a 10-day forecast period. Lower numerical values indicate better performance. }
    \label{fig:single_rmse}
\end{figure}
\subsubsection{Quantitative Skill Evaluation}

We conduct a comparative analysis of PuYun's forecast performance against state-of-the-art ML-based models (Pangu, GraphCast and FuXi) using 0.25° resolution ERA5 data. For 10-day forecasts with a single model, we compared Pangu (using the official greedy strategy across four models), Graphcast, FuXi-Short, and PuYun-Short. Fig. \ref{fig:single_acc} and Fig.\ref{fig:single_rmse} present the globally averaged latitude-weighted ACC and RMSE results for four atmospheric layer variables (Z500, T500, U500, and V500) and four surface variables (MSL, 2T, 10U, and 10V). In terms of the RMSE metric, PuYun-Short outperformed all other models on 10-day. However, in terms of the ACC metric, PuYun-Short slightly lagged behind GraphCast on 2T and T500 on 10-day, while ranking first for all other meteorological variables. We hypothesize that this is due to GraphCast placing greater weight on t2m during training, whereas PuYun assigns equal weight to all meteorological variables. For 10-day forecasts using a single model, PuYun exceeded GraphCast in ACC for 80\% of the meteorological variables, and achieved lower RMSE across all meteorological variables.

For the 10-day forecasts using a model cascading strategy, we primarily focus on the comparison between PuYun and FuXi. Due to the close performance of FuXi and PuYun, the RMSE values are presented in tabular format for clearer comparison. Although the RMSE values for Pangu and Graphcast are also included for reference. The results are shown in Tab. \ref{tab:ca_10days_rmse}.

\begin{table}[htb]
\centering
\caption{10-day forecast comparison across different models.}
\label{tab:ca_10days_rmse}
\begin{tabular}{lcccccccc}
\toprule
 & Z500 & T500 & U500 & V500 & 2T & 10U & 10V & MSL \\ 
\midrule
PanGu         & 789 & 3.36 & 9.03 & 9.28 & 2.78 & 4.5  & 4.68 & 734 \\ 
Graphcast     & 733 & 3.13 & 8.38 & 8.57 & 2.59 & 4.21 & 4.37 & 684 \\ 
FuXi   & 641 & 2.71 & 7.31 & 7.51 & 2.29 & \textbf{3.58} & \textbf{3.72} & \textbf{594} \\ 
PuYun  & \textbf{638} & \textbf{2.68} & \textbf{7.25} & \textbf{7.47} & \textbf{2.28} & 3.59 & 3.73 & 595 \\ 
\bottomrule
\end{tabular}
\end{table}

\subsubsection{Quantitative Predication Evaluation}
\begin{figure}[htbp]
    \centering
    \includegraphics[width=0.93\textwidth]{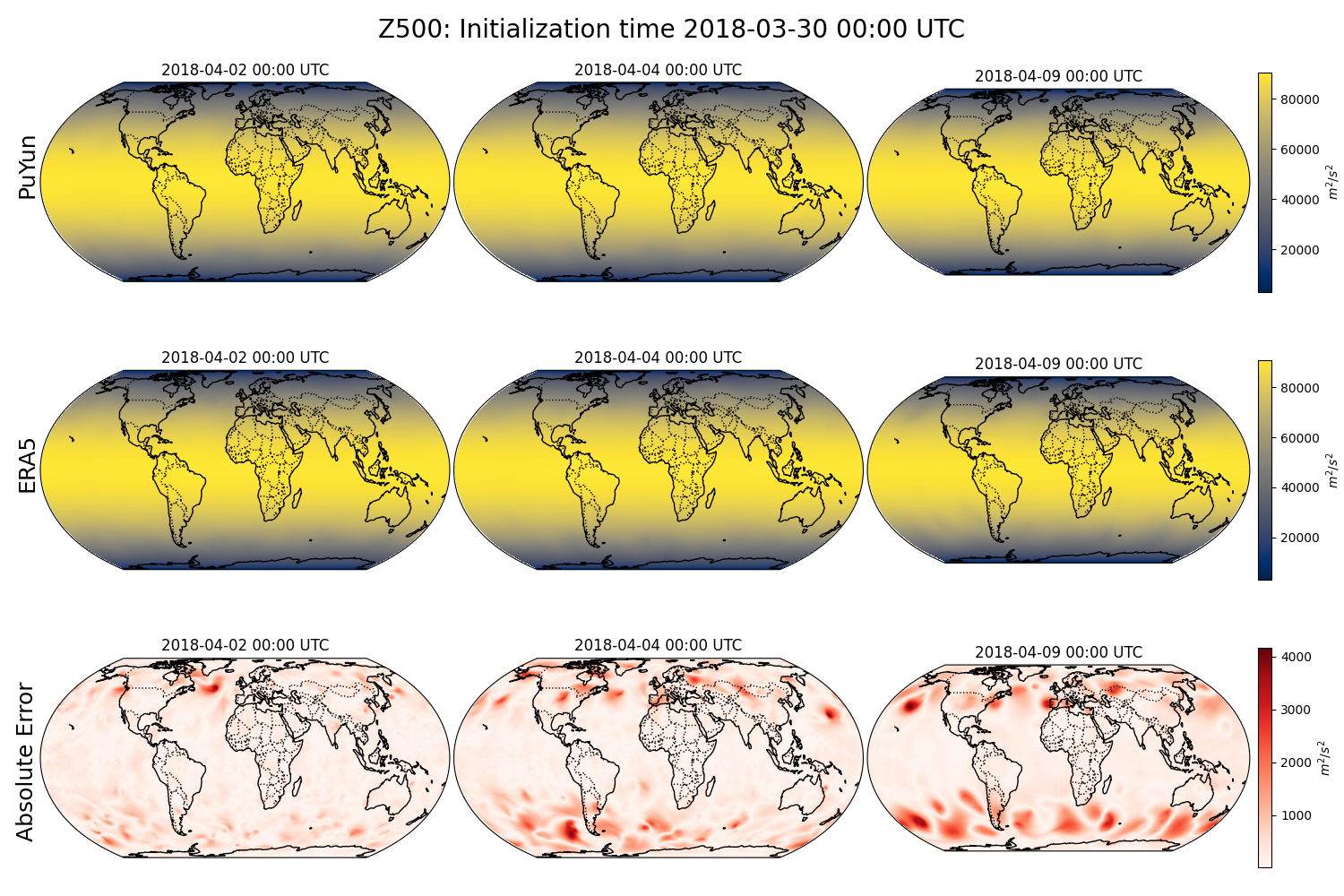}
    \caption{\textbf{Forecast images and absolute error for Z500.} Figures of Z500 on days 3, 5, and 10 are presented with initialization time at 2018-03-30 00:00 UTC.}
    \label{fig:z500}

    \includegraphics[width=0.93\textwidth]{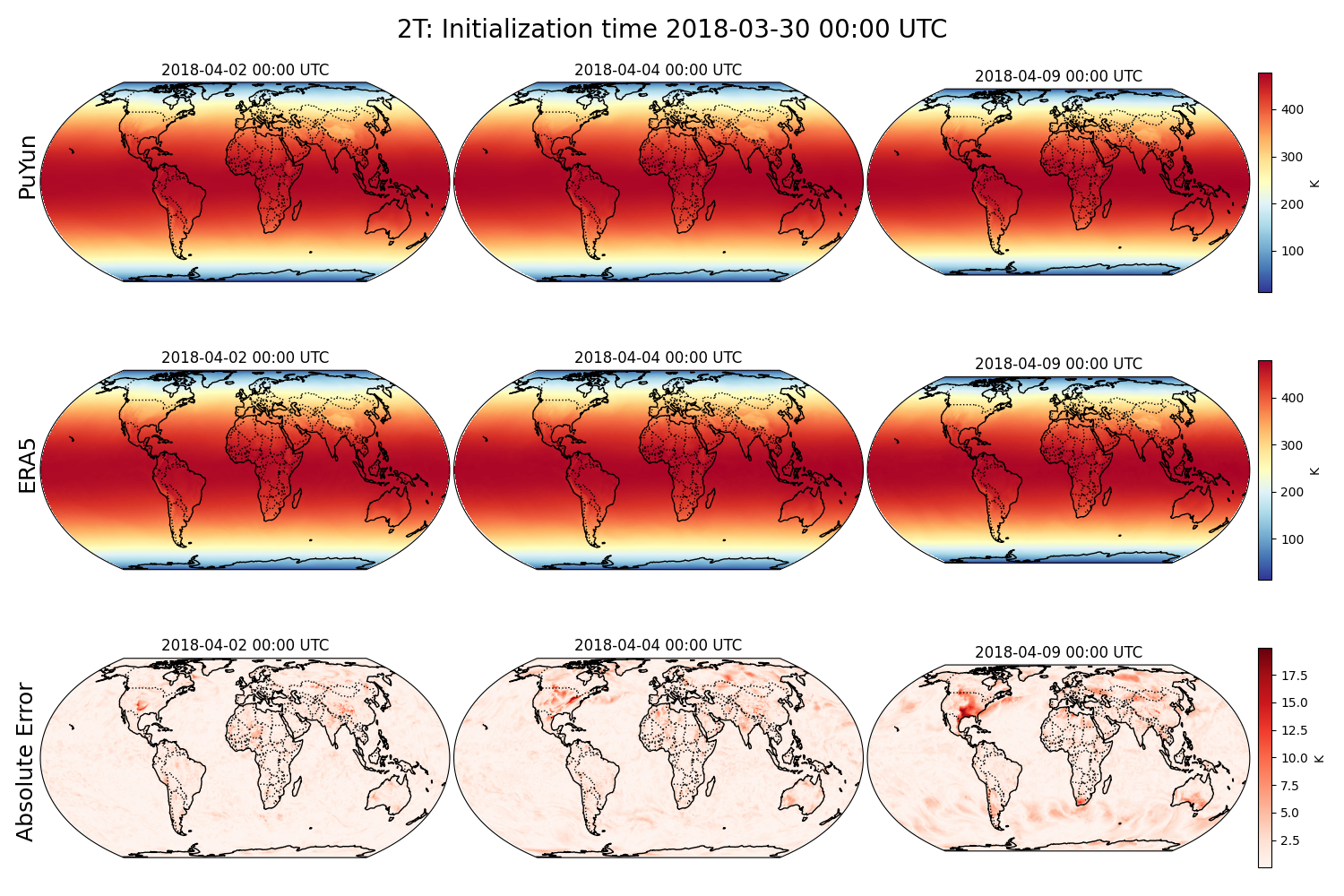}
    \caption{\textbf{Forecast images and absolute error for 2T.} Figures of 2T on days 3, 5, and 10 are presented with initialization time at 2018-03-30 00:00 UTC}
    \label{fig:t2m}
\end{figure}

We present visualizations of PuYun's predicted results at lead days 3, 5, and 10 for two variables: Z500 and 2T, comparing them with ERA5. In Fig. \ref{fig:z500} and Fig. \ref{fig:t2m}, the first two rows display the state sequences from PuYun and ERA5, while the third row shows the absolute error between PuYun and ERA5. The results indicate that PuYun's predictions closely align with ERA5 on the third day. As the forecast lead time increases, the absolute error gradually grows. These visualizations confirm PuYun’s ability to approximate real data and provide accurate weather state estimates.

\subsection{Ablation Experiments}

\begin{table}[t!]
    \centering
    
    \caption{Ablation studies of effectiveness of components for one step prediction.}
    \label{tab:structure-ablation}
    \begin{tabular}{p{7cm}p{1.2cm}p{1.2cm}p{1.2cm}p{1.2cm}}
        \toprule
        Model Arch & Z500 & 2T & 10U & MSL \\
        \midrule
        768d+(6,6,6,6)@K5+[resize] & 26.004 & 0.762 & 0.641 & 36.093 \\
        K7 & 25.545 & 0.748 & 0.629 & 35.442 \\
        K9 & 25.045 & 0.732 & 0.616 & 34.733 \\
        K11 & 25.039 & 0.735 & 0.616 & 34.622 \\
        K9+[pixelshuffle+resize] & 24.813 & 0.724 & 0.560 & 34.416 \\
        (12,12,12,12)@K9+[pixelshuffle+resize] & 23.396 & 0.683 & 0.516 & 32.468 \\
        1536d+(12,12,12,12)@K9+[pixelshuffle+resize] & 22.715 & 0.662 & 0.501 & 30.247 \\
        \bottomrule
    \end{tabular}
    \vskip -3mm
\end{table}

To scrutinize the pivotal components influencing our model's performance, we conducted a series of ablation experiments focusing on the one-step (6h) prediction RMSE values of meteorological elements (Z500, 2T, 10U, MSL) under various structural configurations. These experiments are configured with a default iteration of 30,000 steps, a batch size of 8, and a learning rate of 1e-3, maintaining consistency with all other conditions as described in Section \ref{training}. The first row of Table \ref{tab:structure-ablation} outlines the adjustable structures, including the number of dimensions in the basic block (768d), the number of blocks within each FCN-layer((6,6,6,6)), the kernel size of corresponding convolutions(@K5), and the method of patch merging([resize]). 

Each subsequent row presents a modification relative to the first row's structure, keeping all other aspects unchanged. For instance, rows 2/3/4 only vary in kernel size relative to row 1, whereas row 5 changes both kernel size and the patch merging strategy compared to row 1. The results indicate that larger kernel sizes generally enhance performance, but there is negligible difference between sizes 9 and 11, making size 9 a preferable choice for reducing memory usage. Employing a resize strategy leads to rapid initial convergence, which further improves with the incorporation of the pixelshuffle strategy as the number of steps increases. A substantial reduction in RMSE is observed when the number of dimensions in the basic block is increased from 768 to 1536, and the number of blocks in the FCN-layer is doubled from 6 to 12.

\subsection{Evaluating the Effectiveness of Dynamic Step Autoregressive Training}
After obtaining the model from single-step prediction, we applied dynamic step autoregressive training for fine-tuning to enhance the model's performance in multi-step forecasting. In dynamic step autoregressive training, a maximum step length M is set, and during each training iteration, different GPUs are randomly assigned autoregressive steps ranging from 2 to M. At the end of each iteration, the gradients from all GPUs are aggregated, followed by a model parameter update.

\begin{figure}[htbp]
    \centering
    \vspace{-20pt}
    \includegraphics[width=1.0\textwidth]{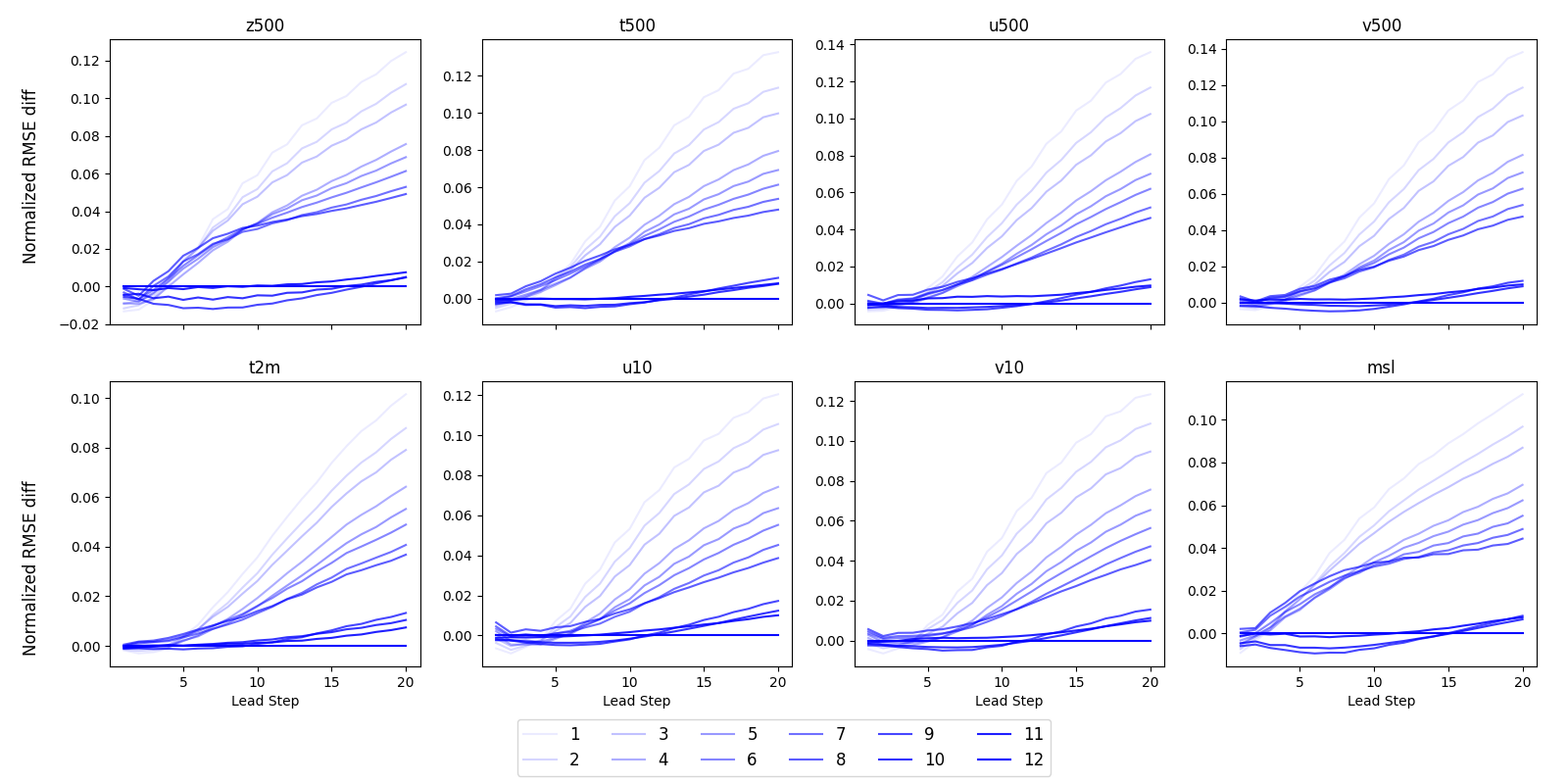}
    \caption{\textbf{The effectiveness of dynamic step autoregressive training.} Evaluate the impact of the maximum regression steps on the results using normalized globally-averaged latitude-weighted RMSE using testing data from 2018.}
    \label{fig:norm}

    \includegraphics[width=1.0\textwidth]{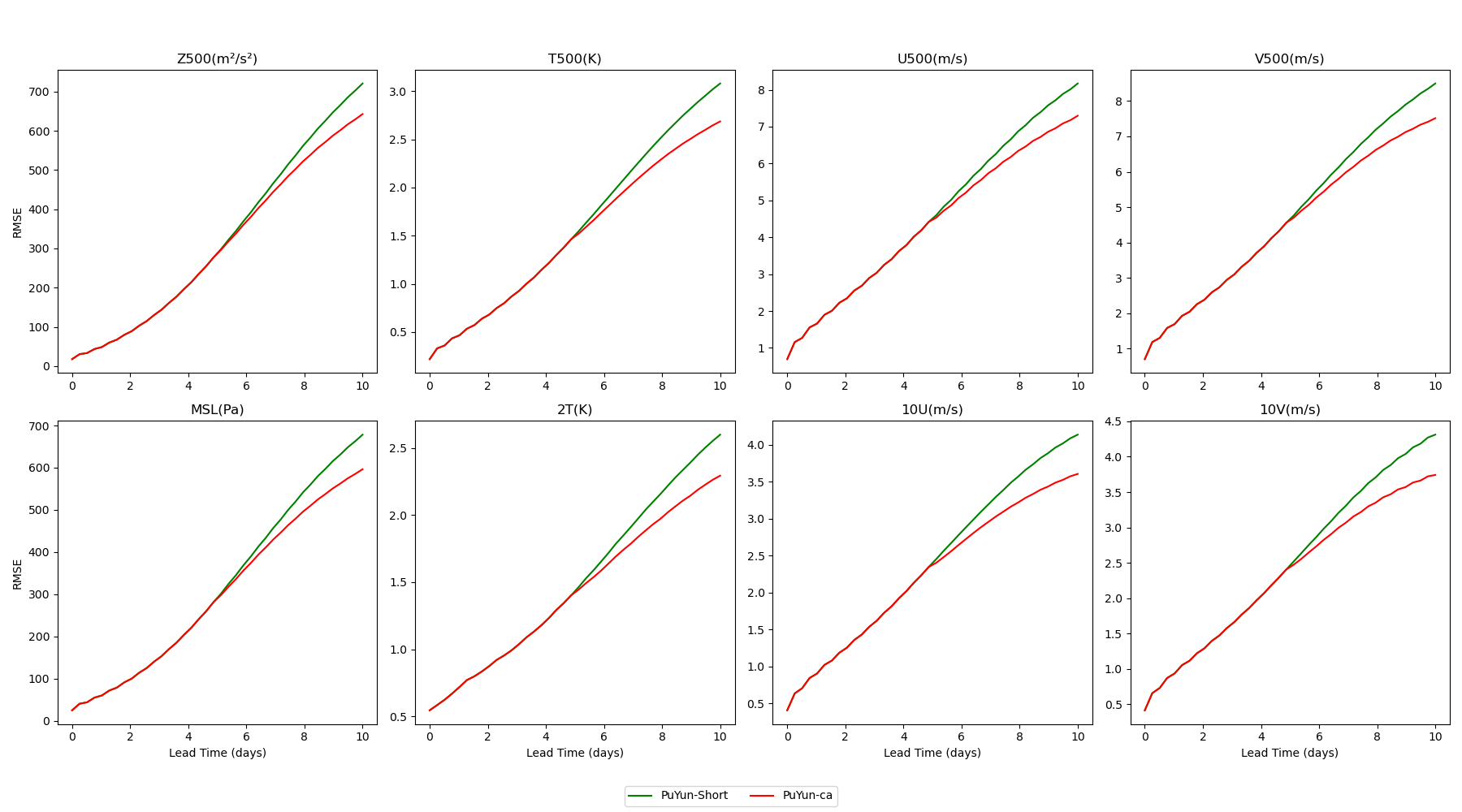}
    \caption{\textbf{The impact of cascading.} Comparison of the globally-averaged latitude-weighted RMSE of the PuYun and PuYun without cascade using testing data from 2018.}
    \label{fig:ca}
    
\end{figure}

We gradually increased M from 2 to 12, training for 3000 iterations at each experiment. The results, as shown in the Fig. \ref{fig:norm}, indicate that larger M lead to lower RMSE at longer prediction steps. Specifically, when M=12, the RMSE for 20-step predictions is more than 10\% lower compared to M=1.

\subsection{Effectiveness of the Cascade Model Design}
In this section, we ablate over the influence of the cascade ML model architecture on alleviating accumulation errors in weather forecasting. We utilize a single PuYun model (PuYun-Short) to generate 10-day forecasts and evaluate its efficacy in comparison to PuYun (A cascade model of PuYun-Short and PuYun-Medium). Fig. \ref{fig:ca} illustrates the performance PuYun-Short and PuYun, spanning lead times from 0 to 10 days. The performance of single PuYun-Short predictions diminishes significantly with increasing lead times. We assert this is primarily attributed to accumulation errors.

The experimental results of the cascade ML model architecture reveal its efficacy in mitigating accumulation errors in weather forecasting. Despite the success observed in the initial lead times, as depicted in Fig. \ref{fig:ca}, a single model is shown to be incapable of achieving optimal forecast performance across various lead times.

\section{Conclusion}
In recent years, data-driven ML models have made significant advancements in global medium-range weather forecasting, surpassing even the HRES. However, these ML models, while resource-intensive during training, are currently confined to a spatial resolution of 0.25°. Such resolution falls significantly short of practical operational demands in critical industries like renewable energy generation, agriculture, and transportation. Furthermore, the persistent challenge of accumulation errors in medium-range forecasts adds another layer of complexity to accurate predictions.

In response to these challenges, we introduce PuYun, a novel approach based on LKA-FCN featuring a cascading structure. PuYun excels in generating 10-day forecasts, updating every 6 hours, encompassing four surface variables and five atmospheric variables across 13 vertical pressure levels.

Leveraging LKA convolutions enhances PuYun's effective receptive field, allowing it to capture meteorological changes in adjacent areas and gain a deeper understanding of the meteorological system. PuYun introduces dynamic step autoregressive training, extending the model's predictive capability from single-step to multi-step forecasts. Using only a single model, PuYun-Short achieved state-of-the-art performance in predicting global weather for the next 10 days, highlighting the potential of dynamic step autoregressive training and large kernel attention (LKA) in meteorological forecasting. When employing a model cascading strategy for forecasting global weather over the next 10 days, PuYun also achieved world-leading performance on over 80\% of meteorological variables. This result further underscores the effectiveness of the model cascading approach.

PuYun demonstrates the capability to be fine-tuned on the HRES-fc0-0.25d dataset\footnote{The HRES-fc0 dataset refers to the initial time step of ECMWF's High-Resolution Ensemble Forecast (HRES) predictions. The HRES-fc0 dataset also encompasses resolutions of 0.1° and 0.25°, denoted as HRES-fc0-0.1d and HRES-fc0-0.25d, respectively.} and subsequently utilize the HRES-fc0-0.1d dataset as input for generating global forecasts at a higher resolution of 0.1°. This capability highlights our forthcoming objective of achieving high-resolution global predictions.

\section{Future Work}
It is essential to recognize that current machine learning models heavily rely on initial fields produced by numerical weather prediction (NWP) models. We are witnessing an exponential increase in global meteorological observation instruments, accompanied by a continual enhancement in the temporal resolution of observations. Looking ahead, our objective is to develop end-to-end global weather forecasting algorithms that utilize global observational data as input, enabling real-time generation of medium-range weather forecasts, while also advancing the forecasting resolution to 0.1°.

\par\vfill\par

\bibliographystyle{splncs04}
\bibliography{main}  
\clearpage






\end{document}